\documentclass[10pt,twocolumn,letterpaper]{article}

\usepackage{cvpr}
\usepackage{threeparttable}
\usepackage{bbding}
\usepackage[percent]{overpic}
\usepackage[dvipsnames]{xcolor}
\usepackage{times}
\usepackage{epsfig}
\usepackage{graphicx}
\usepackage{amsmath}
\usepackage{amssymb}
\usepackage{mathtools}
\usepackage{multirow}
\usepackage{enumitem}
\usepackage{caption}
\usepackage{subcaption}
\usepackage{array}
\usepackage{colortbl}
\usepackage{booktabs}
\usepackage{bbm}
\usepackage{multicol}
\usepackage{makecell}
\usepackage{xspace}
\usepackage{float}
\usepackage{color}
\usepackage{lipsum}
\usepackage{pifont}
\usepackage{cite}

\usepackage[pagebackref=true,breaklinks=true,letterpaper=true,colorlinks,bookmarks=false]{hyperref}
\usepackage[capitalize]{cleveref}

\usepackage{array}
\newcolumntype{H}{>{\setbox0=\hbox\bgroup}c<{\egroup}@{}}

\cvprfinalcopy 

\def\cvprPaperID{****} 

\newcommand\blfootnote[1]{%
  \begingroup
  \renewcommand\thefootnote{}\footnote{#1}%
  \addtocounter{footnote}{-1}%
  \endgroup
}
\begin{document}

\title{YOLOv6 v3.0: A Full-Scale Reloading}

\author{Chuyi Li$^*$~~~ Lulu Li$^*$~~~ Yifei Geng$^*$~~~Hongliang Jiang$^*$\\MengCheng~~~ Bo Zhang~~~ Zaidan Ke~~~ Xiaoming Xu$^\dag$~~~ Xiangxiang Chu \\
	Meituan Inc.\\
	\tt\small \{lichuyi, lilulu05, gengyifei, jianghongliang02,\\
  \tt\small   chengmeng05, zhangbo97, kezaidan, xuxiaoming04, chuxiangxiang\}@meituan.com \\
    }

\twocolumn[
{\renewcommand\twocolumn[1][]{#1}
\maketitle
\vspace{-11mm}
\begin{figure}[H]
\hsize=\textwidth
\centering
\begin{subfigure}{0.48\textwidth}
\centering
\includegraphics[width=1\textwidth]{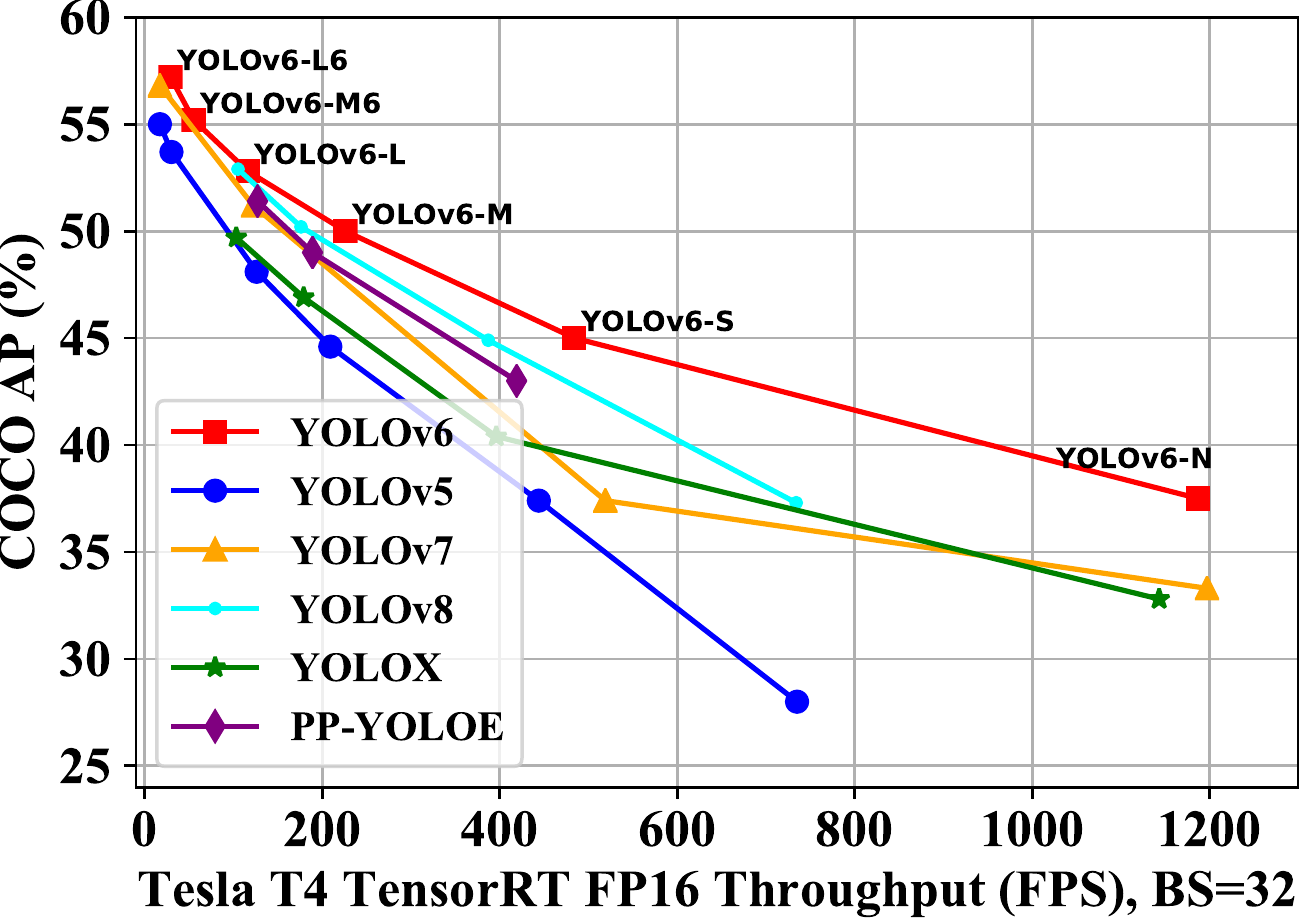}
\label{fig:1a}	
\end{subfigure}    
\hspace{0.2in}
\begin{subfigure}{0.48\textwidth}
\centering
\includegraphics[width=1\textwidth]{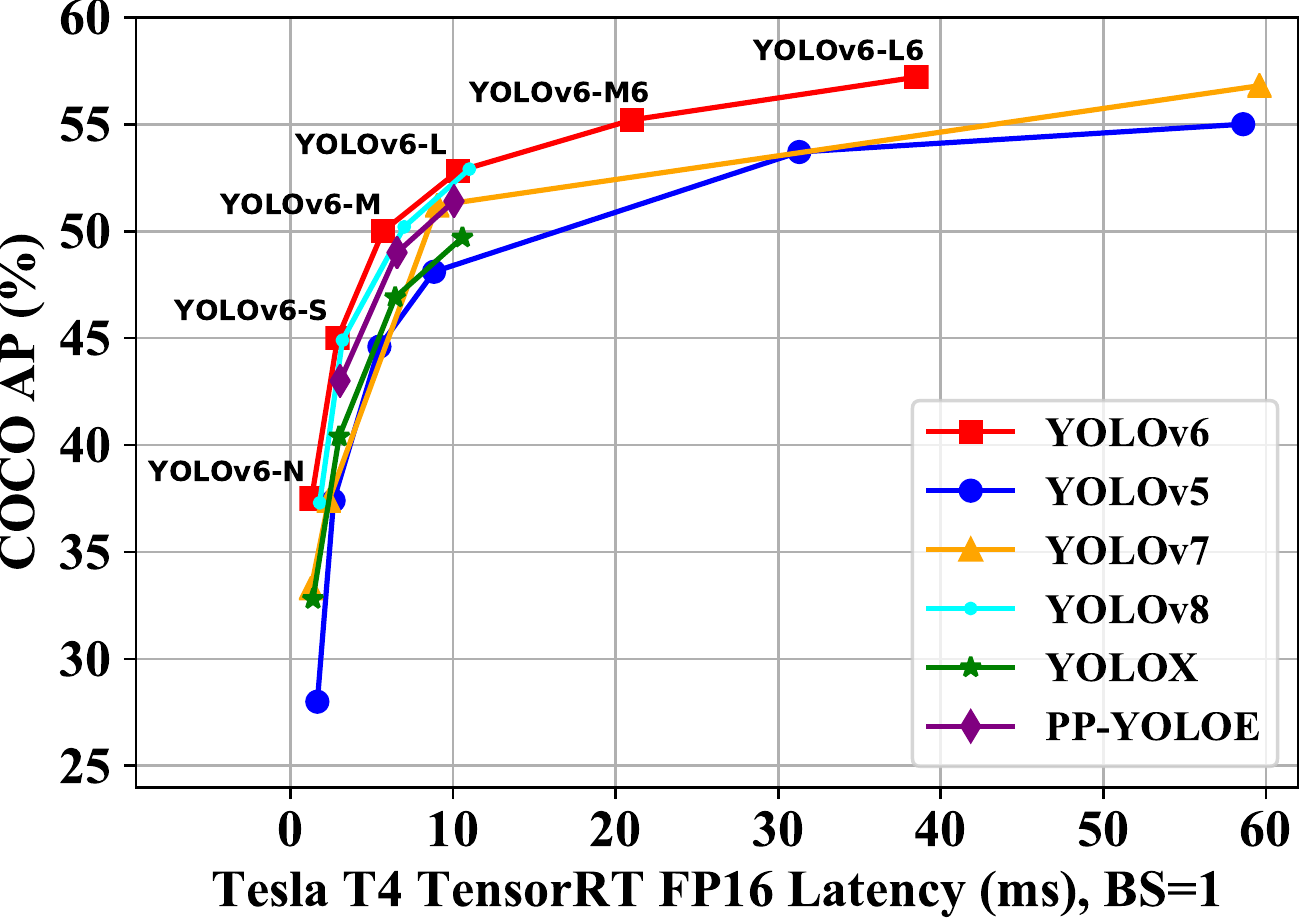}
\label{fig:1b}
\end{subfigure}
\hspace{0.in}
\vspace{-6mm}
\caption{Comparison of state-of-the-art efficient object detectors. Both latency and throughput (at a batch size of 32) are given for a handy reference. All models are test with TensorRT 7.}
\label{fig:sota-comp}
\end{figure}
}
]

\blfootnote{* Equal contributions.}
\blfootnote{\dag~Corresponding author.}

\begin{abstract}
  The YOLO community has been in high spirits since our first two releases! By the advent of  Chinese New Year 2023,  which sees the Year of the Rabbit, we refurnish YOLOv6 with numerous novel  enhancements on the network architecture and the training scheme. This release is identified as YOLOv6 v3.0. For a glimpse of performance, our YOLOv6-N hits 37.5\% AP on the COCO dataset at a throughput of 1187 FPS tested with an NVIDIA Tesla T4 GPU. YOLOv6-S strikes 45.0\% AP at 484 FPS, outperforming other mainstream detectors at the same scale~(YOLOv5-S, YOLOv8-S, YOLOX-S and PPYOLOE-S). Whereas, YOLOv6-M/L also achieve better accuracy performance (50.0\%/52.8\% respectively) than other detectors at a similar inference speed. Additionally, with an extended backbone and neck design, our YOLOv6-L6 achieves the state-of-the-art accuracy in real-time. Extensive experiments are carefully conducted to validate the effectiveness of each improving component. Our code is made available at~\url{https://github.com/meituan/YOLOv6}.
\end{abstract}


\begin{figure*}[t]
  \begin{center}
    \includegraphics[width=0.99\linewidth]{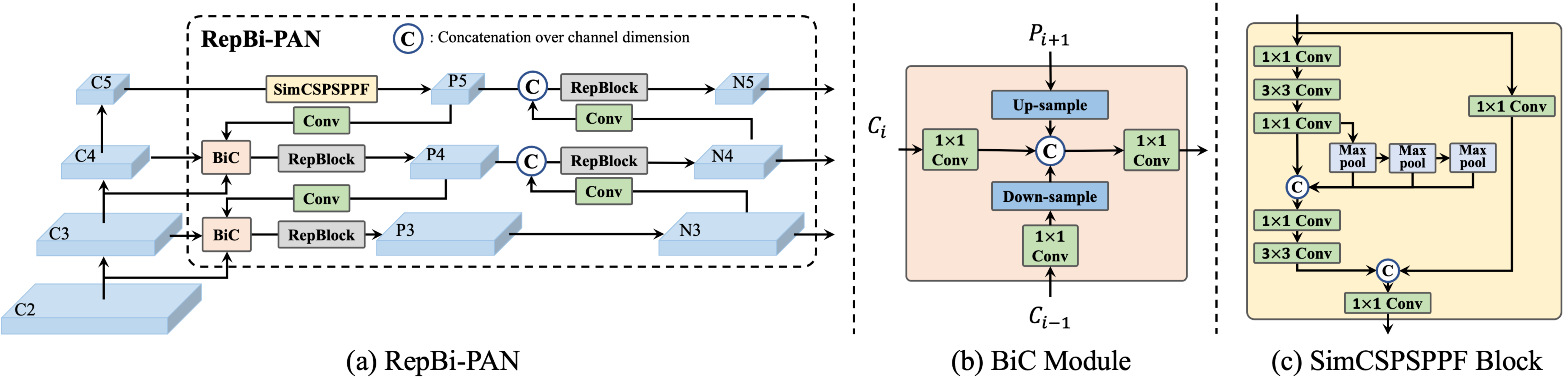}
  \end{center}
    \vskip -0.2in
    \caption{(a) The neck of YOLOv6 (N and S are shown). Note for M/L, RepBlocks is replaced with CSPStackRep. (b) The structure of a BiC module. (c) A SimCSPSPPF block.}
    \label{fig:neck}
\end{figure*}

\section{Introduction}
\label{sec:intro}
The YOLO series has been the most popular detection frameworks in industrial applications, for its excellent balance between speed and accuracy. Pioneering works of YOLO series are YOLOv1-3~\cite{redmon2016you,redmon2017yolo9000,redmon2018yolov3}, which blaze a new trail of one-stage detectors along with the later substantial improvements. YOLOv4~\cite{bochkovskiy2020yolov4} reorganized the detection framework into several separate parts (backbone, neck and head), and verified bag-of-freebies and bag-of-specials at the time to design a framework suitable for training on a single GPU. At present, YOLOv5~\cite{yolov5}, YOLOX~\cite{ge2021yolox}, PPYOLOE~\cite{xu2022ppyoloe}, YOLOv7~\cite{wang2022yolov7} and most recently YOLOv8~\cite{yolov8} are all the competing candidates for efficient detectors to deploy. 

In this release, we strenuously renovate the network design and the training strategy. We show the comparison of YOLOv6 with other peers at a similar scale in~\cref{fig:sota-comp}. The new features of YOLOv6 are summarized as follows:

\begin{itemize}
  \item We renew the neck of the detector with a Bi-directional Concatenation (BiC) module to provide more accurate localization signals. SPPF~\cite{yolov5} is simplified to form the SimCSPSPPF Block, which brings performance gains with negligible speed degradation.
  
  \item We propose an \emph{anchor-aided training} (AAT) strategy to enjoy the advantages of both anchor-based and anchor-free paradigms without touching inference efficiency.
  
  \item We deepen YOLOv6 to have another stage in the backbone and the neck, which reinforces it to hit a new state-of-the-art performance on the COCO dataset at a high-resolution input.

  \item We involve a new self-distillation strategy to boost the performance of small models of YOLOv6, in which the heavier branch for DFL~\cite{li2020generalized} is taken as an \emph{enhanced auxiliary regression branch} during training and is removed at inference to avoid the marked speed decline.
\end{itemize}

\section{Method}
\label{sec:method}

\subsection{Network Design}
  \label{sec:method:network}
  In practice, feature integration at multiple scales has been proven to be a critical and effective component of object detection. Feature Pyramid Network (FPN)~\cite{lin2017feature} is proposed to aggregate the high-level semantic features and low-level features via a top-down pathway, which provides more accurate localization. Subsequently, there have been several works~\cite{liu2018path, tan2020efficientdet, ghiasi2019fpn, chen2021parallel} on Bi-directional FPN in order to enhance the ability of hierarchical feature representation. PANet~\cite{liu2018path} adds an extra bottom-up pathway on top of FPN to shorten the information path of low-level and top-level features, which facilitates the propagation of accurate signals from low-level features. BiFPN~\cite{tan2020efficientdet} introduces learnable weights for different input features and simplifies PAN to achieve better performance with high efficiency. PRB-FPN \cite{chen2021parallel} is proposed to retain high-quality features for accurate localization by a parallel FP structure with bi-directional fusion and associated improvements.

  Motivated by the above works, we design an enhanced-PAN as our detection neck. In order to augment localization signals without bringing in excessive computation burden, we propose a Bi-directional Concatenation(BiC) module to integrate feature maps of three adjacent layers, which fuses an extra low-level feature from backbone $C_{i-1}$ into $P_i$ (Fig.~\ref{fig:neck}). In this case, more accurate localization signals can be preserved, which is significant for the localization of small objects. 

  Moreover, we simplify the SPPF block \cite{yolov5} to have a CSP-like version called SimCSPSPPF Block, which strengthens the representational ability. Particularly, we revise the SimSPPCSPC Block in~\cite{wang2022yolov7} by shrinking the channels of hidden layers and retouching space pyramid pooling. In addition, we upgrade the CSPBlock with RepBlock (for small models) or CSPStackRep Block (for large models) and accordingly adjust the width and depth. The neck of YOLOv6 is denoted as RepBi-PAN, the framework of which is shown in~\cref{fig:neck}.

\subsection{Anchor-Aided Training}

  YOLOv6 is an anchor-free detector to pursue a higher inference speed. However, we experimentally find that the anchor-based paradigm brings additional performance gains on YOLOv6-N under the same settings when compared with the anchor-free one, as shown in~\cref{tab:comp-parad}. Moreover, anchor-based ATSS~\cite{zhang2020atss} is adopted as the warm-up label assignment strategy in the early versions of YOLOv6, which stabilizes the training.
  
  \begin{table}[ht]
  \centering
  \scalebox{0.9}{
    \begin{tabular}{l|cccc}
      \toprule
      \textbf{Paradigm} & \textbf{AP$^{val}$} & \textbf{AP$^{s}$} & \textbf{AP$^{m}$} & \textbf{AP$^{l}$} \\
      \midrule
      \midrule
      Anchor-free & 35.5\% &16.0\% &39.5\% & 51.0\% \\
      Anchor-based & \textbf{35.6}\%  &\textbf{17.2}\%  &\textbf{39.8}\%& \textbf{52.1}\%  \\
      \bottomrule
    \end{tabular}
  }
  \caption{Comparisons of the anchor-free and the anchor-based paradigms on YOLOv6-N.}
  \label{tab:comp-parad}
\end{table}

\begin{figure}[htp]
  \centering
  \includegraphics[width=\columnwidth]{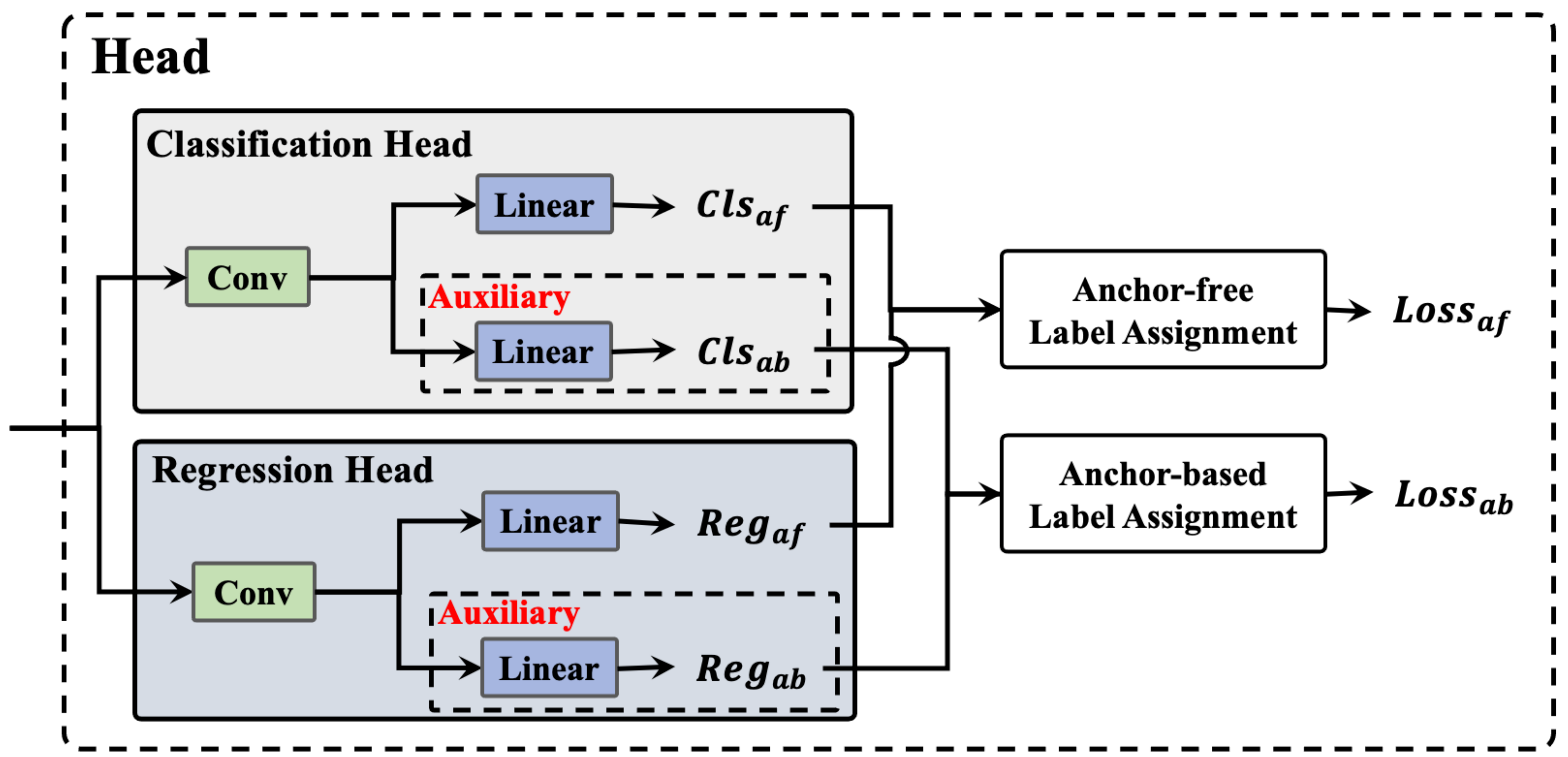}
  \caption{The detection head with anchor-based auxiliary branches during training. The auxiliary branches are removed at inference. `af' and `ab' are short for `anchor-free' and `anchor-based'. }
  \label{fig:auod}
\end{figure}

  In light of this, we propose \emph{anchor-aided training (AAT)}, in which the \emph{anchor-based auxiliary branches} are introduced to combine the advantages of anchor-based and anchor-free paradigms. And they are applied both in the classification and the regression head. \cref{fig:auod} shows the detection head with the auxiliaries.

  During the training stage, the auxiliary branches and the anchor-free branches learn from independent losses while signals are propagated altogether. Therefore, additional embedded guidance information from auxiliary branches is integrated into the main anchor-free heads. Worth mentioning that the auxiliary branches is removed at inference, which boosts the accuracy performance without decreasing speed.
  
  \subsection{Self-distillation}
    In early versions of YOLOv6, the self-distillation is only introduced in large models (i.e., YOLOv6-M/L), which applies the vanilla knowledge distillation technique by minimizing the KL-divergence between the class prediction of the teacher and the student. Meanwhile DFL~\cite{li2020generalized} is adopted as regression loss to perform self-distillation on box regression similar to~\cite{zheng2022localization}.

  The knowledge distillation loss is formulated as: 
 \begin{equation}
 L_{KD} = KL(p_t^{cls}||p_s^{cls}) + KL(p_t^{reg}||p_s^{reg}),
 \end{equation}
 where $p_t^{cls}$ and $p_s^{cls}$ are class prediction of the teacher model and the student model respectively, and accordingly $p_t^{reg}$ and $p_s^{reg}$ are box regression predictions. The overall loss function is now formulated as:
 \begin{equation}
 L_{total} = L_{det} + \alpha L_{KD},
 \end{equation}
 where $L_{det}$ is the detection loss computed with predictions and labels. The hyperparameter $\alpha$ is introduced to balance two losses. In the early stage of training, the soft labels from the teacher are easier to learn. As the training continues, the performance of the student will match the teacher so that the hard labels will help students more. Upon this, we apply \emph{cosine weight decay} to $\alpha$ to dynamically adjust the information from hard labels and soft ones from the teacher. 
  The formulation of $\alpha$ is:
 \begin{equation}
  \alpha = -0.99 * ((1-cos(\pi * E_i / E_{max})) / 2) + 1,
 \end{equation}
where $E_i$ denotes the current training epoch and $E_{max}$ represents the maximum training epochs.
 
Notably, the introduction of DFL~\cite{li2020generalized} requires extra parameters for the regression branch, which affects the inference speed of small models significantly. Therefore, we specifically design the \emph{Decoupled Localization Distillation} (DLD) for our small models to boost performance without speed degradation. Specifically, we append a heavy auxiliary \emph{enhanced regression branch} to incorporate DFL. During the self-distillation, the student is equipped with a na\"ive regression branch and the enhanced regression branch while the teacher only uses the auxiliary branch. Note that the na\"ive regression branch is only trained with hard labels while the auxiliary is updated according to signals from both the teacher and hard labels. After the distillation, the na\"ive regression branch is retained whilst the auxiliary branch is removed. With this strategy, the advantages of the heavy regression branch for DFL in distillation is considerably maintained without impacting the inference efficiency.
  
  \begin{table*}[h]
    \centering
    \resizebox{0.85\textwidth}{!}{
      \begin{tabular}{l|c|c|c|c|c|c|c|c}
        \toprule
        \multirow{2}{*}{\textbf{Method}} & \multirow{2}{*}{\textbf{Input Size}} & \multirow{2}{*}{\textbf{AP}$^{val}$} & \multirow{2}{*}{\textbf{AP}$_{50}^{val}$} &\multirow{2}{*}{\textbf{FPS}} &\multirow{2}{*}{\textbf{FPS}} &\multirow{2}{*}{\textbf{Latency}} & \multirow{2}{*}{\textbf{Params}} & \multirow{2}{*}{\textbf{FLOPs}} \\
        & & & & \tiny{(bs=1)} & \tiny{(bs=32)} &\tiny{(bs=1)} & & \\			
        \midrule
        \midrule
        YOLOv5-N~\cite{yolov5} & 640 & 28.0\% & 45.7\% &602  & 735 & 1.7 ms& 1.9 M & 4.5 G\\
        YOLOv5-S~\cite{yolov5} & 640 & 37.4\% & 56.8\% & 376 & 444 & 2.7 ms& 7.2 M & 16.5 G \\
        YOLOv5-M~\cite{yolov5} & 640 & 45.4\% & 64.1\% & 182 & 209 & 5.5 ms& 21.2 M & 49.0 G  \\
        YOLOv5-L~\cite{yolov5} & 640 & 49.0\% & 67.3\% &113  & 126 & 8.8 ms& 46.5 M & 109.1 G  \\
        \midrule
        YOLOv5-N6~\cite{yolov5} & 1280 &36.0\% & 54.4\%&172 &175 &5.8 ms &3.2 M &18.4 G \\
        YOLOv5-S6~\cite{yolov5} & 1280 &44.8\% & 63.7\%&103 &103 &9.7 ms &12.6 M &67.2 G \\
        YOLOv5-M6~\cite{yolov5} & 1280 &51.3\% & 69.3\%&49 &48 &20.1 ms &35.7 M &200.0 G \\
        YOLOv5-L6~\cite{yolov5} & 1280 &53.7\% & 71.3\%&32 &30 &31.3 ms &76.8 M &445.6 G \\
        YOLOv5-X6~\cite{yolov5} & 1280 &55.0\% & 72.7\%&17 &17 &58.6 ms &140.7 M &839.2 G \\
        \midrule
        \midrule
        YOLOX-Tiny~\cite{ge2021yolox} & 416 & 32.8\% & 50.3\%$^*$  &717  & 1143 &1.4 ms & 5.1 M & 6.5 G \\
        YOLOX-S~\cite{ge2021yolox} & 640 & 40.5\% & 59.3\%$^*$ &333  & 396 & 3.0 ms & 9.0 M & 26.8 G \\
        YOLOX-M~\cite{ge2021yolox} & 640 & 46.9\% & 65.6\%$^*$ &155  & 179 & 6.4 ms &25.3 M & 73.8 G \\
        YOLOX-L~\cite{ge2021yolox} & 640 & 49.7\% & 68.0\%$^*$ &94  & 103 & 10.6 ms &54.2 M &155.6 G \\
        \midrule
        \midrule
        PPYOLOE-S~\cite{xu2022ppyoloe} & 640 & 43.1\% & 59.6\% & 327 & 419 & 3.1 ms& 7.9 M & 17.4 G \\
        PPYOLOE-M~\cite{xu2022ppyoloe} & 640 & 49.0\% & 65.9\% & 152 & 189 & 6.6 ms& 23.4 M& 49.9 G \\
        PPYOLOE-L~\cite{xu2022ppyoloe} & 640 & 51.4\% & 68.6\% & 101 & 127 & 10.1 ms& 52.2 M& 110.1 G \\
        \midrule
        \midrule
        YOLOv7-Tiny~\cite{wang2022yolov7} & 416 &33.3\%$^*$ & 49.9\%$^*$ & 787  &1196  & 1.3 ms & 6.2 M & 5.8 G \\
        YOLOv7-Tiny~\cite{wang2022yolov7} & 640 &37.4\%$^*$ & 55.2\%$^*$ & 424  & 519 & 2.4 ms & 6.2 M & 13.7 G$^*$ \\
        YOLOv7~\cite{wang2022yolov7}  & 640 & 51.2\% & 69.7\%$^*$ & 110 & 122 & 9.0 ms & 36.9 M& 104.7 G \\
        YOLOv7-E6E~\cite{wang2022yolov7}  & 1280 & 56.8\% & 74.4\%$^*$ & 16 & 17 & 59.6 ms & 151.7 M& 843.2 G \\
        \midrule
        \midrule
        YOLOv8-N~\cite{yolov8} & 640 & 37.3\% & 52.6\%$^*$ & 561  & 734 & 1.8 ms& 3.2 M & 8.7 G\\
        YOLOv8-S~\cite{yolov8} & 640 & 44.9\% & 61.8\%$^*$ & 311 & 387 & 3.2 ms& 11.2 M & 28.6 G \\
        YOLOv8-M~\cite{yolov8} & 640 & 50.2\% & 67.2\%$^*$ & 143 & 176 & 7.0 ms& 25.9 M & 78.9 G  \\
        YOLOv8-L~\cite{yolov8} & 640 & 52.9\% & 69.8\%$^*$ & 91  & 105 & 11.0 ms& 43.7 M & 165.2 G  \\

        \midrule
        \midrule
        YOLOv6-N & 640 &  37.0\% / 37.5\%$^\ddagger$ & 52.7\% / 53.1\%$^\ddagger$ & 779 & 1187 &1.3 ms &4.7 M & 11.4 G \\
        YOLOv6-S & 640 & 44.3\% / 45.0\%$^\ddagger$ & 61.2\% / 61.8\%$^\ddagger$ & 339 & 484 &2.9 ms & 18.5 M & 45.3 G \\
        YOLOv6-M & 640 & 49.1\% / 50.0\%$^\ddagger$ & 66.1\% / 66.9\%$^\ddagger$ & 175 & 226 & 5.7 ms & 34.9 M & 85.8 G \\
        YOLOv6-L & 640 & 51.8\% / 52.8\%$^\ddagger$ & 69.2\% / 70.3\%$^\ddagger$ & 98  & 116 &10.3 ms & 59.6 M & 150.7 G \\
        
        \midrule
        YOLOv6-N6 & 1280 & 44.9\% & 61.5\% &228 &281 & 4.4 ms &10.4 M &49.8 G\\
        YOLOv6-S6 & 1280 & 50.3\% &67.7\% & 98& 108&10.2 ms &41.4 M &198.0 G\\
        YOLOv6-M6 & 1280 & 55.2\%$^\ddagger$ &72.4\%$^\ddagger$ &47 & 55&21.0 ms &79.6 M &379.5 G\\
        YOLOv6-L6 & 1280 & 57.2\%$^\ddagger$ &74.5\%$^\ddagger$  & 26 & 29 & 38.5 ms & 140.4 M & 673.4 G\\

        \bottomrule
      \end{tabular}
    }
    \caption{
      Comparisons with other YOLO-series detectors on COCO 2017 \emph{val}. FPS and latency are measured  in FP16-precision on a Tesla T4 in the same environment with TensorRT. All our models are trained for 300 epochs without pre-training or any external data. Both the accuracy and the speed performance of our models are evaluated with the input resolution of 640$\times$640. `$\ddagger$' represents that the proposed self-distillation method is utilized. `$*$' represents the re-evaluated result of the released model through the official code.
    }
    \label{tab:sota}
  \end{table*}

\section{Experiments}
\label{sec:exp}

\subsection{Comparisons}
  
  The evaluation is made consistent with the early versions of YOLOv6~\cite{li2022yolov6}, which focuses on the throughput  and the GPU latency at deployment. We test the speed performance of all official models with FP16-precision on the same Tesla T4 GPU with TensorRT~\cite{tensorrt}. We compare the upgraded YOLOv6 with YOLOv5~\cite{yolov5}, YOLOX~\cite{ge2021yolox}, PPYOLOE~\cite{xu2022ppyoloe}, YOLOv7~\cite{wang2022yolov7} and YOLOv8~\cite{yolov8}. Note that the performance of YOLOv7-Tiny is re-evaluated according to their open-sourced code and weights at the input size of 416 and 640. Results are shown in~\cref{tab:sota} and~\cref{fig:sota-comp}.
  Compared with YOLOv5-N/YOLOv7-Tiny (input size=416), our YOLOv6-N has significantly advanced by 9.5\%/4.2\% respectively. It also comes with the best speed performance in terms of both throughput and latency. Compared with YOLOX-S/PPYOLOE-S, YOLOv6-S can improve AP by 3.5\%/0.9\% with higher speed. 
  YOLOv6-M outperforms YOLOv5-M by 4.6\% higher AP with a similar speed, and it achieves 3.1\%/1.0\% higher AP than YOLOX-M/PPYOLOE-M at a higher speed. Besides, it is more accurate and faster than YOLOv5-L. YOLOv6-L is 3.1\%/1.4\% more accurate than YOLOX-L/PPYOLOE-L under the same latency constraint. Compared with the YOLOv8 series, our YOLOv6 achieves a similar performance in accuracy and in the latency for models at all sizes, while giving significant better throughput performance.

  To compare with the state-of-the-art methods, we follow~\cite{yolov5} to add an extra stage on the top of the backbone to have a feature (C6) at a higher level for detecting extra-large objects. The neck is also expanded accordingly. The  YOLOv6 of all sizes with C6 features are named YOLOv6-N6/S6/M6/L6 respectively. 
  Further, the image resolution is adapted from 640 to 1280. The feature strides range from 8 to 64, which benefits the accurate detection for rather small and extra-large objects in high-resolution images. The experimental results listed in~\cref{tab:sota} show that the expanded YOLOv6 obtain significant gains in accuracy. Compared with expanded YOLOv5 (i.e., YOLOv5-N6/S6/M6/L6/X6), ours have a much higher AP at a similar inference speed. When compared with the state-of-the-art YOLOv7-E6E, YOLOv6-L6 improves AP by 0.4\% and runs 63\% faster with a batch size of 1.

  \begin{table}[ht]
    \centering
    \scalebox{0.9}{
      \begin{tabular}{ccc|c}
        \toprule
        \textbf{BiC+SimCSPSPPF} & \textbf{AAT} & \textbf{DLD} &\textbf{AP$^{val}$} \\
        \midrule
        \midrule
          \XSolidBrush & \XSolidBrush & \XSolidBrush & 43.5\%\\
          \Checkmark & \XSolidBrush  & \XSolidBrush & 44.1\% \\
          \Checkmark  & \Checkmark   & \XSolidBrush & 44.4\% \\
          \Checkmark  & \Checkmark & \Checkmark & 45.1\%  \\

        \bottomrule
      \end{tabular}
    }
  \caption{Ablation study for all designs on YOLOv6-S.}
  \label{tab:ablate:total}
  \end{table}

  \begin{table}[!thbp]
    \begin{center}
      \scalebox{0.45}{
        \resizebox{\textwidth}{!}{
            \begin{tabular}{l|cc|ccccc}
              \toprule
              \multirow{2}{*}{\textbf{Model}} & \multicolumn{2}{c}{\textbf{BiC}} & \multirow{2}{*}{\textbf{AP$^{val}$}} &\multirow{2}{*}{\textbf{AP$^{s}$}} & \multirow{2}{*}{\textbf{AP$^{m}$}} & \multirow{2}{*}{\textbf{AP$^{l}$}} & \multirow{2}{*}{\textbf{FPS}} \\
               & \textbf{Bottom-up} & \textbf{Top-down} & & & & &  \tiny{(bs=32)}  \\
              \midrule
              \midrule
              \multirow{3}{*}{YOLOv6-S} 
                & \XSolidBrush & \XSolidBrush & 43.1\% & 23.4\% & 48.0\% & 59.9\% & \textbf{513} \\
                & \XSolidBrush & \Checkmark & \textbf{43.7}\% & \textbf{25.2}\% & \textbf{48.7}\% & \textbf{60.4}\% & 492 \\
                & \Checkmark & \Checkmark & \textbf{43.7}\% & 25.0\% & \textbf{48.7}\% & 59.7\% & 485 \\
              \midrule
              \multirow{3}{*}{YOLOv6-L} 
                & \XSolidBrush & \XSolidBrush & 50.9\% & 32.4\% & 56.0\% & \textbf{68.0\%} & \textbf{125} \\
                & \XSolidBrush & \Checkmark & \textbf{51.3}\% & \textbf{34.2}\% & 56.5\% & 67.6\% & 120 \\
                & \Checkmark & \Checkmark & 51.1\% & 33.6\% & \textbf{56.7}\% & 67.9\% & 119 \\
            \bottomrule
            \end{tabular}
      }}
    \end{center}
    \caption{Effectiveness of the BiC module on YOLOv6.}
    \label{tab:ablate:neck:bic}
  \end{table}

\subsection{Ablation Study}
Experimental results in~\cref{tab:ablate:total} exhibit the effectiveness of all contributions in this work. The renovated network with BiC and SimCSPSPPF has an enhanced AP by 0.6\%. With AAT and DLD, the accuracy is further improved by 0.3\% and 0.7\% incrementally.

\subsubsection{Network Design}

  We conduct a series of experiments to verify the effectiveness of the proposed BiC module. As can be seen in~\cref{tab:ablate:neck:bic}, applying the BiC module only on the top-down pathway of PAN brings 0.6\%/0.4\% AP improvements on YOLOv6-S/L respectively with negligible loss of efficiency. In contrast, when we try to import the BiC module into the bottom-up pathway, no positive gain in accuracy is obtained. The probable reason is that the BiC module on the bottom-up pathway would lead to confusion for detection heads about features at different scales. Therefore, we merely adopt the BiC module on the top-down pathway. Besides, the results indicate that the BiC module gives an impressive boost to the performance of small object detection. For both YOLOv6-S and YOLOv6-L, the detection performance on small objects is improved by 1.8\%.

  Further, we explore the influence of different types of SPP Blocks, including the simplified variants of SPPF~\cite{yolov5} and SPPCSPC~\cite{wang2022yolov7} (denoted as SimSPPF and SimSPPCSPC respectively) and our SimCSPSPPF blocks. Additionally, we apply SimSPPF blocks on the top three feature maps (P3, P4 and P5) of our backbone to verify its effectiveness, which is denoted as SimSPPF*3. Experimental results are shown in~\cref{tab:ablate:neck:spp}. We observe that heavily adopting SimSPPF brings little gain in accuracy with the increased computational complexity. SimSPPCSPC outperforms SimSPPF by 1.6\%/0.3\% AP on YOLOv6-N/S respectively while significantly decreasing inference speed. Compared with SimSPPF, our SimCSPSPPF version can obtain 1.1\%/0.4\%/0.1\% performance gain for YOLOv6-N/S/M respectively. In terms of inference efficiency, our SimCSPSPPF block runs nearly 10\% faster than SimSPPCSPC and is slightly slower than SimSPPF. For a better accuracy-efficiency trade-off, the SimCSPSPPF blocks are introduced in YOLOv6-N/S. For YOLOv6-M/L, SimSPPF blocks are adopted.

  \begin{table}[ht]
    \centering
    \scalebox{0.9}{
      \begin{tabular}{l|c|ccc}
        \toprule
        \multirow{2}{*}{\textbf{Model}} & \multirow{2}{*}{\textbf{SPP Blocks}} & \multirow{2}{*}{\textbf{AP$^{val}$}} & \multirow{2}{*}{\textbf{FPS}} \\
          & & &  \tiny{(bs=32)} \\
        \midrule
        \midrule
        \multirow{4}{*}{YOLOv6-N} 
          & SimSPPF~\cite{yolov5} & 35.8\% & \textbf{1190}  \\
          & SimSPPF~\cite{yolov5}*3 & 35.9\% & 1072  \\
          & SimSPPCSPC~\cite{wang2022yolov7} & \textbf{37.4}\%  &1078  \\
          & SimCSPSPPF & 36.9\%  & 1176  \\
        \midrule
        \multirow{4}{*}{YOLOv6-S} 
          & SimSPPF~\cite{yolov5} & 43.7\% &  \textbf{492}  \\
          & SimSPPF~\cite{yolov5}*3 & 43.6\%  &447  \\
          & SimSPPCSPC~\cite{wang2022yolov7} &44.0\%  &432  \\
          & SimCSPSPPF & \textbf{44.1}\%  &477  \\
        \midrule
        \multirow{2}{*}{YOLOv6-M} 
          & SimSPPF~\cite{yolov5} & 48.6\% &  \textbf{227}  \\
          & SimCSPSPPF & \textbf{48.7}\%  &218  \\
        \midrule
        \multirow{2}{*}{YOLOv6-L} 
          & SimSPPF~\cite{yolov5} & \textbf{51.3}\% &  \textbf{120}  \\
          & SimCSPSPPF & 51.1\%  &117  \\
        \bottomrule
      \end{tabular}
    }
  \caption{Ablation study on different types of SPP Blocks. All models are equipped with BiC modules.}
  \label{tab:ablate:neck:spp}
  \end{table}

  \begin{table}
    \centering
    \scalebox{0.7}{
      \begin{tabular}{l|c|cccc}
        \toprule
        \textbf{Method} & \textbf{AAT} & \textbf{AP$^{val}$} & \textbf{AP$^{s}$} & \textbf{AP$^{m}$} & \textbf{AP$^{l}$}\\
        \midrule
        \midrule
        \multirow{2}{*}{YOLOv6-N}
        & \XSolidBrush & \textbf{36.9}\% &17.2\% &41.1\% &52.9\% \\
        & \Checkmark & \textbf{36.9}\% & \textbf{18.7}\% & \textbf{41.2}\% & \textbf{53.0}\% \\
        \midrule
        \multirow{2}{*}{YOLOv6-S}
        & \XSolidBrush &44.1\% &24.7\% &48.7\% & \textbf{61.1}\% \\
        & \Checkmark & \textbf{44.4}\%  &\textbf{25.4}\% &\textbf{49.6}\% &60.2\% \\
        \midrule
        \multirow{2}{*}{YOLOv6-M}
        & \XSolidBrush &48.6\% &29.7\% &53.7\% &\textbf{65.5}\%  \\
        & \Checkmark & \textbf{49.1}\% & \textbf{31.1}\% & \textbf{54.0}\% &65.4\% \\
        \midrule
        \multirow{2}{*}{YOLOv6-L}
        & \XSolidBrush & 51.3\% &\textbf{34.2}\% &56.5\% &67.6\%  \\
        & \Checkmark & \textbf{51.8}\% &33.4\% &\textbf{56.8}\% &\textbf{68.8}\% \\
        \bottomrule
      \end{tabular}
    }
    \caption{Ablation study about AAT.}
    \label{tab:ablate:head}
  \end{table}

  \subsubsection{Anchor-Aided Training}
    The advantages of the AAT is verified in YOLOv6. As shown in~\cref{tab:ablate:head}, it brings about 0.3\%/0.5\%/0.5\% AP gain for YOLOv6-S/M/L respectively. Notably, the accuracy performance on small objects ($AP^{s}$) is significantly enhanced for YOLOv6-N/S/M. For YOLOv6-L, the performance on large objects ($AP^{l}$) is improved even further.

\subsubsection{Self-distillation}

\begin{table}[ht]
  \centering
  \scalebox{0.9}{
    \begin{tabular}{l|c|c}
      \toprule
      \textbf{Model} & \textbf{Weight Decay} & \textbf{AP$^{val}$} \\
      \midrule
      \midrule
      Baseline & - & 51.8\% \\
      \midrule
      Double epochs & - & 51.7\% \\
      \midrule
      \multirow{2}{*}{Self-distillation}
      &\XSolidBrush &51.8\%  \\
      &\Checkmark &\textbf{52.4}\%  \\
      \bottomrule
    \end{tabular}
  }
  \caption{Ablation on the self-distillation on YOLOv6-L.}
  \label{tab:distillation}
\end{table}

\begin{table}
  \centering
  \scalebox{0.9}{
    \begin{tabular}{cc|c}
      \toprule
      \textbf{DLD} & \textbf{Double epochs}& \textbf{AP$^{val}$} \\
      \midrule
      \midrule
      \XSolidBrush&\XSolidBrush & 44.4\% \\
      \XSolidBrush & \Checkmark & 44.6\% \\
      \Checkmark & \XSolidBrush & \textbf{45.1}\% \\
      \bottomrule
    \end{tabular}
  }
  \caption{Ablation study of DLD on YOLOv6-S.}
  \label{tab:dld}
\end{table}

We verify the proposed self-distillation method on YOLOv6-L. For a fair comparison, we also verified the model performance by doubling the training epochs besides the baseline since the self-distillation needs an extra entire training cycle to obtain the teacher model. As seen in~\cref{tab:distillation}, no performance improvement is attained without the weight decay strategy compared with the baseline. Doubling the training epochs is even worse due to overfitting. After the introduction of weight decay, the model is boosted by 0.6\% AP.

In addition, the DLD specifically designed for small models is ablated on YOLOv6-S. As per self-distillation for large models, we also compare the results with the model trained with doubled epochs. As shown in~\cref{tab:dld}, YOLOv6-S with DLD gives 0.7\% AP boost and performs 0.5\% better than that of training with doubled epochs.

\section{Conclusion}
In this report, YOLOv6 is upgraded on aspects of the network design and training strategy, which boost  YOLOv6 to achieve the state-of-the-art accuracy for real-time object detection. In the future, we persistently work on the optimization of YOLOv6 to render an application-friendly detection framework as our research continues and the techniques in object detection advance.

{\small
\bibliographystyle{ieee_fullname}
\bibliography{egbib}
}

\clearpage

\appendix

\section{Detailed Latency and Throughput Benchmark}\label{app:bench}
\subsection{Setup}
Unless otherwise stated, all the reported latency is measured on an NVIDIA Tesla T4 GPU with TensorRT version 7.2.1.6. Due to the large variance of the hardware and software settings, we re-measure latency and throughput of all the models under the same configuration (both hardware and software). For a handy reference, we also switch TensorRT versions (Table~\ref{tab:latency-qps-trt82}) for consistency check. Latency on a V100 GPU (Table~\ref{tab:latency-qps-v100}) is included for a convenient comparison. This gives us a full spectrum view of state-of-the-art detectors.

\subsection{T4 GPU Latency Table with TensorRT 8}

See \cref{tab:latency-qps-trt82}. The throughput of YOLOv6 models still emulates their peers.

\begin{table}[ht]
	\centering
	\resizebox{0.8\columnwidth}{!}{
		\begin{tabular}{l|HHHc|c|cHH}
			\toprule
			\multirow{2}{*}{\textbf{Method}} & \multirow{2}{*}{\textbf{Input}} & \multirow{2}{*}{\textbf{AP}$^{val}$} & \multirow{2}{*}{\textbf{AP}$_{50}^{val}$} &\multirow{2}{*}{\textbf{FPS}} &\multirow{2}{*}{\textbf{FPS}} &\multirow{2}{*}{\textbf{Latency}} & \multirow{2}{*}{\textbf{Params}} & \multirow{2}{*}{\textbf{FLOPs}} \\
			& & & & \tiny{(bs=1)} & \tiny{(bs=32)} &\tiny{(bs=1)} & & \\			
			\midrule
			\midrule
      YOLOv5-N~\cite{yolov5} & 640 & 28.0\% & 45.7\% & 702 & 843 & 1.4 ms& 1.9 M & 4.5 G\\
      YOLOv5-S~\cite{yolov5} & 640 & 37.4\% & 56.8\% & 433 & 515 & 2.3 ms& 7.2 M & 16.5 G \\
      YOLOv5-M~\cite{yolov5} & 640 & 45.4\% & 64.1\% & 202 & 235 & 4.9 ms& 21.2 M & 49.0 G  \\
      YOLOv5-L~\cite{yolov5} & 640 & 49.0\% & 67.3\% & 126  & 137 & 7.9 ms& 46.5 M & 109.1 G  \\
      \midrule
			\midrule
      YOLOX-Tiny~\cite{ge2021yolox} & 416 & 32.8\% & 50.3\%$^*$  & 766 & 1393 &1.3 ms & 5.1 M & 6.5 G \\
      YOLOX-S~\cite{ge2021yolox} & 640 & 40.5\% & 59.3\%$^*$ & 313  & 489 & 2.6 ms & 9.0 M & 26.8 G \\
      YOLOX-M~\cite{ge2021yolox} & 640 & 46.9\% & 65.6\%$^*$ & 159  & 204 & 5.3 ms &25.3 M & 73.8 G \\
      YOLOX-L~\cite{ge2021yolox} & 640 & 49.7\% & 68.0\%$^*$ &104  & 117 & 9.0 ms &54.2 M &155.6 G \\
      \midrule
      \midrule
      PPYOLOE-S~\cite{xu2022ppyoloe} & 640 & 43.1\% & 59.6\% & 357 & 493 & 2.8 ms& 7.9 M & 17.4 G \\
      PPYOLOE-M~\cite{xu2022ppyoloe} & 640 & 49.0\% & 65.9\% & 163 & 210 & 6.1 ms& 23.4 M& 49.9 G \\
      PPYOLOE-L~\cite{xu2022ppyoloe} & 640 & 51.4\% & 68.6\% & 110  & 145 & 9.1 ms& 52.2 M& 110.1 G \\
      \midrule
      \midrule
      YOLOv7-Tiny~\cite{wang2022yolov7} & 640 &37.4\%$^*$ & 55.2\%$^*$ & 464  &   568 & 2.1 ms & 6.2 M & 13.7 G$^*$ \\
      YOLOv7~\cite{wang2022yolov7}  & 640 & 51.2\% & 69.7\% & 128 & 135 & 7.6 ms & 36.9 M& 104.7 G \\
      \midrule
      \midrule
      YOLOv6-N & 640 &  37.0\% / 37.5\%$^\ddagger$ & 52.7\% / 53.1\%$^\ddagger$ & 785 & 1215 &1.3 ms &4.7 M & 11.4 G \\
      YOLOv6-S & 640 & 44.3\% / 45.0\%$^\ddagger$ & 61.2\% / 61.8\%$^\ddagger$ & 345 & 498 &2.9 ms & 18.5 M & 45.3 G \\
      YOLOv6-M & 640 & 49.1\% / 50.0\%$^\ddagger$ & 66.1\% / 66.9\%$^\ddagger$ & 178 & 238 & 5.6 ms & 34.9 M & 85.8 G \\
      YOLOv6-L & 640 & 51.8\% / 52.8\%$^\ddagger$ & 69.2\% / 70.3\%$^\ddagger$ & 105 & 125 & 9.5 ms & 59.6 M & 150.7 G \\
      \bottomrule
		\end{tabular}
	}
	\caption{
    YOLO-series comparison of latency and throughput on a T4 GPU with a higher version of TensorRT (8.2).
	}
	\label{tab:latency-qps-trt82}
\end{table}

\subsection{V100 GPU Latency Table}
See~\cref{tab:latency-qps-v100}. The speed advantage of YOLOv6 is largely maintained.

\begin{table}[ht]
	\centering
	\resizebox{0.8\columnwidth}{!}{
		\begin{tabular}{l|HHHc|c|cHH}
			\toprule
			\multirow{2}{*}{\textbf{Method}} & \multirow{2}{*}{\textbf{Input}} & \multirow{2}{*}{\textbf{AP}$^{val}$} & \multirow{2}{*}{\textbf{AP}$_{50}^{val}$} &\multirow{2}{*}{\textbf{FPS}} &\multirow{2}{*}{\textbf{FPS}} &\multirow{2}{*}{\textbf{Latency}} & \multirow{2}{*}{\textbf{Params}} & \multirow{2}{*}{\textbf{FLOPs}} \\
			& & & & \tiny{(bs=1)} & \tiny{(bs=32)} &\tiny{(bs=1)} & & \\			
			\midrule
			\midrule
      YOLOv5-N~\cite{yolov5} & 640 & 28.0\% & 45.7\% & 577  & 1727 & 1.4 ms& 1.9 M & 4.5 G\\
      YOLOv5-S~\cite{yolov5} & 640 & 37.4\% & 56.8\% & 449 & 1249 & 1.7 ms& 7.2 M & 16.5 G \\
      YOLOv5-M~\cite{yolov5} & 640 & 45.4\% & 64.1\% & 271 & 698 & 3.0 ms& 21.2 M & 49.0 G  \\
      YOLOv5-L~\cite{yolov5} & 640 & 49.0\% & 67.3\% & 178  & 440 & 4.7 ms& 46.5 M & 109.1 G  \\
      \midrule
			\midrule
      YOLOX-Tiny~\cite{ge2021yolox} & 416 & 32.8\% & 50.3\%$^*$  & 569  & 2883 &1.4 ms & 5.1 M & 6.5 G \\
      YOLOX-S~\cite{ge2021yolox} & 640 & 40.5\% & 59.3\%$^*$ &386  & 1206 & 2.0 ms & 9.0 M & 26.8 G \\
      YOLOX-M~\cite{ge2021yolox} & 640 & 46.9\% & 65.6\%$^*$ &245  & 600 & 3.4 ms &25.3 M & 73.8 G \\
      YOLOX-L~\cite{ge2021yolox} & 640 & 49.7\% & 68.0\%$^*$ &149  & 361 & 5.6 ms &54.2 M &155.6 G \\
      \midrule
      \midrule
      PPYOLOE-S~\cite{xu2022ppyoloe} & 640 & 43.1\% & 59.6\% & 322 & 1050 & 2.4 ms& 7.9 M & 17.4 G \\
      PPYOLOE-M~\cite{xu2022ppyoloe} & 640 & 49.0\% & 65.9\% & 222 & 566 & 4.0 ms& 23.4 M& 49.9 G \\
      PPYOLOE-L~\cite{xu2022ppyoloe} & 640 & 51.4\% & 68.6\% & 153 & 406 & 5.5 ms& 52.2 M& 110.1 G \\
      \midrule
      \midrule
      YOLOv7-Tiny~\cite{wang2022yolov7} & 640 &37.4\%$^*$ & 55.2\%$^*$ & 453  & 1565 & 1.7 ms & 6.2 M & 13.7 G$^*$ \\
      YOLOv7~\cite{wang2022yolov7}  & 640 & 51.2\% & 69.7\% & 182 & 412 & 4.6 ms & 36.9 M& 104.7 G \\
      \midrule
      \midrule
      YOLOv6-N & 640 &  37.0\% / 37.5\%$^\ddagger$ & 52.7\% / 53.1\%$^\ddagger$ & 646 & 2660 &1.2 ms &4.7 M & 11.4 G \\
      YOLOv6-S & 640 & 44.3\% / 45.0\%$^\ddagger$ & 61.2\% / 61.8\%$^\ddagger$ & 399 & 1330 &2.0 ms & 22.4 M & 45.3 G \\
      YOLOv6-M & 640 & 49.1\% / 50.0\%$^\ddagger$ & 66.1\% / 66.9\%$^\ddagger$ & 203 & 676 & 4.4 ms & 34.9 M & 85.8 G \\
      YOLOv6-L & 640 & 51.8\% / 52.8\%$^\ddagger$ & 69.2\% / 70.3\%$^\ddagger$ & 125 & 385 & 6.8 ms & 59.6 M & 150.7 G \\
      \bottomrule
		\end{tabular}
	}
	\caption{
    YOLO-series comparison of latency and throughput on a V100 GPU. We measure all models at FP16-precision with the input size 640$\times$640 in the exact same environment.
	}
	\label{tab:latency-qps-v100}
\end{table}

\subsection{CPU Latency}
We evaluate the performance of our models and other competitors on a 2.6 GHz  Intel Core i7 CPU using OpenCV Deep Neural Network (DNN), as shown in~\cref{tab:latency-cpu}.

\begin{table}[ht]
	\centering
	\resizebox{0.7\columnwidth}{!}{
		\begin{tabular}{l|cHHHHcHH}
			\toprule
			\multirow{2}{*}{\textbf{Method}} & \multirow{2}{*}{\textbf{Input}} & \multirow{2}{*}{\textbf{AP}$^{val}$} & \multirow{2}{*}{\textbf{AP}$_{50}^{val}$} &\multirow{2}{*}{\textbf{FPS}} &\multirow{2}{*}{\textbf{FPS}} &\multirow{2}{*}{\textbf{Latency}} & \multirow{2}{*}{\textbf{Params}} & \multirow{2}{*}{\textbf{FLOPs}} \\
			& & & & \tiny{(bs=1)} & \tiny{(bs=32)} &\tiny{(bs=1)} & & \\			
			\midrule
			\midrule
      YOLOv5-N~\cite{yolov5} & 640 & 28.0\% & 45.7\% &691  & 1199 & 118.9 ms& 1.9 M & 4.5 G\\
      YOLOv5-S~\cite{yolov5} & 640 & 37.4\% & 56.8\% & 552 & 1054 & 202.2 ms& 7.2 M & 16.5 G \\
      \midrule
			\midrule
      YOLOX-Tiny~\cite{ge2021yolox} & 416 & 32.8\% & 50.3\%$^*$  &738  & 3074 &144.2 ms & 5.1 M & 6.5 G \\
      YOLOX-S~\cite{ge2021yolox} & 640 & 40.5\% & 59.3\%$^*$ &499  & 1223 & 164.6 ms & 9.0 M & 26.8 G \\
      YOLOX-M~\cite{ge2021yolox} & 640 & 46.9\% & 65.6\%$^*$ &295  & 598 & 357.9 ms &25.3 M & 73.8 G \\
      \midrule
      \midrule
      YOLOv7-Tiny~\cite{wang2022yolov7} & 640 &37.4\%$^*$ & 55.2\%$^*$ & 527  & 697 & 137.5 ms & 6.2 M & 13.7 G$^*$ \\
      \midrule
      \midrule
      YOLOv6-N & 640 & 35.9\% & 51.2\% & 752 & 2031 &60.3 ms &4.3 M & 11.1 G \\
      YOLOv6-S & 640 & 40.3\% & 56.6\% & 604 & 1724 & 148.0 ms & 15.0 M & 36.7 G \\
      YOLOv6-M & 640 & 43.5\% & 60.4\% & 507 & 1546 &269.3 ms & 17.2 M & 44.2 G \\
      \bottomrule
		\end{tabular}
	}
	\caption{
    YOLO-series comparison of latency on a typical CPU. We measure all models at FP32-precision with the input size 640$\times$640 in the exact same environment.
	}
	\label{tab:latency-cpu}
\end{table}

\end{document}